\pdfoutput=1
\documentclass[letterpaper, 10 pt, conference]{ieeeconf}  % Comment this line out if you need a4paper

\IEEEoverridecommandlockouts                              % This command is only needed if 
\usepackage{mathptmx} % assumes new font selection scheme installed
\usepackage{amsmath} % assumes amsmath package installed
\usepackage{amssymb}  % assumes amsmath package installed
\usepackage{graphicx}
\usepackage{subcaption}
\long\def\invis#1{}
\usepackage{cite}
\usepackage{url}
\usepackage{hyphenat}
\usepackage{amsmath,systeme}
\usepackage{bm}
\usepackage{xcolor}
\usepackage[linesnumbered,ruled]{algorithm2e}
\usepackage{siunitx}
\usepackage{enumerate}
\usepackage{mathtools}
\usepackage{booktabs}

\usepackage[utf8]{inputenc}
\usepackage{ wasysym }
\usepackage{pgfplots}
\usepackage{tikz}
\usepackage{gensymb}
\usepackage{pgfplotstable}
\usepgfplotslibrary{colormaps}
\pgfplotsset{colormap/hot}
\pgfplotsset{width=9cm,compat=1.9}
\pgfplotsset{layers/my layer set/.define layer set={background,main,foreground}{ },set layers=my layer set,}
\usepackage{xcolor}
\usepackage{tabularx}
\usepackage{multirow}
\definecolor{darkblue}{rgb}{0.0, 0.0, 0.55}
\usepackage[colorlinks=true, urlcolor=darkblue]{hyperref}

\newcommand\fig[1]{Figure~\ref{#1}}

\newcommand\bs{\boldsymbol}
\newcommand\bmat{\begin{bmatrix}}
\newcommand\emat{\end{bmatrix}}
\usepackage{textcomp}
\usepackage{svg}
\usepackage{algpseudocode}
\usepackage{hyperref}
\usepackage{makecell}
\usepackage{soul}

\def\floatsep{4.0pt}
\def\textfloatsep{0.0pt}
\def\dblfloatsep{4.0pt}
\def\dbltextfloatsep{4.0pt}

\usepackage{mathtools}
\DeclarePairedDelimiterX{\norm}[1]{\lVert}{\rVert}{#1}

\newcommand{\R}[2]{{}^{#2}_{#1}\mathbf{R}}
\newcommand{\Rhat}[2]{{}^{#2}_{#1}\hat{\mathbf{R}}}

\newcommand{\quat}[2]{{}^{#2}_{#1}\bar{q}}

\newcommand{\p}[2]{{}^{#2}\mathbf{p}_{#1}}
\newcommand{\phat}[2]{{}^{#2}\hat{\mathbf{p}}_{#1}}
\newcommand{\ptilde}[2]{{}^{#2}\tilde{\mathbf{p}}_{#1}}
\newcommand{\vel}[2]{{}^{#2}\mathbf{v}_{#1}}

\DeclareMathSymbol{\shortminus}{\mathbin}{AMSa}{"39}

\usepackage{fp}
\FPset{\pb}{0}
\newcommand{\pagebudget}[1]{}

\usepackage{todonotes}

\usepackage{xcolor}

\newif\ifmarkchange
\definecolor{changed}{rgb}{0.0,0.0,0.0}
\ifmarkchange
\definecolor{changed}{rgb}{0.91, 0.05, 0.05}
\fi

\title{
\LARGE \bf SVIn+: Multi-Modal Framework for Robust Underwater State Estimation using 3D Sonar, Visual-Inertial, Water-Pressure, and DVL}
\author{Chinmay Burgul$^1$, Xiaoming Zhao$^2$, Yewei Huang$^2$, Michalis Chatzispyrou$^1$, \\ \hspace{0.5cm} Marios Xanthidis$^3$, Ioannis Rekleitis$^1$, Alberto Quattrini Li$^2$
\thanks{$^1$ Chinmay Burgul, Michalis Chatzispyrou, and Ioannis Rekleitis are with  the Mechanical Engineering Department, University of Delaware, Newark, DE, USA. {\tt\small [cmburgul,michalis,yiannisr]@udel.edu}}
\thanks{$^2$ Xiaoming Zhao, Yewei Huang and Alberto Quattrini Li are with Dartmouth College, Hanover, NH, USA {\tt\small \{yewei.huang, alberto.quattrini.li\}@dartmouth.edu}}
\thanks{$^3$ Marios Xanthidis is with the Aquaculture Robotics and Automation Group, SINTEF Ocean, {\tt\small marios.xanthidis@sintef.no}}
\thanks{This work was supported by the Research Council of Norway (EchoNav: NO-359447), National Science Foundation (NSF 2545370, 1943205, 2024541, 2144624)}
}

\begin{document}

\maketitle
\thispagestyle{empty}
\pagestyle{empty}

%%%%%%%%%%%%%%%%%%%%%%%%%%%%%%%%%%%
%%%%%%%%%%%%%%%%%%%%%%%%%%%%%%%%%%%
%%%%%%%%%%%%%%%%%%%%%%%%%%%%%%%%%%%

% \begin{IEEEkeywords}
% \end{IEEEkeywords}
% \IEEEpeerreviewmaketitle
%%%%%%%%%%%%%%%%%%%%%%%%%%%%%%%%%%%%%%%%%%%%%%%%%%%%%%%%%%%%%%%%%%%%%%%%%%%%%%%%
\begin{abstract}
% संदर्भ
As expansion and scalability in underwater operations becomes more important and challenging, autonomous underwater vehicles (AUVs) become more relevant and the necessity for reliable autonomy more critical.
% समस्या
A fundamental challenge in any autonomous underwater operation, from control to safe navigation and mapping, is state estimation; a notorious hard problem to address, especially in low visibility conditions.
% Central to achieving this is robust state estimation -- particularly in low visibility environments -- which serves as the fundamental prerequisite for safe navigation, motion planning and control.
% उपाय
In this paper, we tackle this problem, by presenting SVIn+, a multi-sensor framework for robust state estimation of AUVs equipped with an IMU, cameras, a water-pressure sensor, a Doppler Velocity Log (DVL), and a compact 3D sonar.
This framework extends an established open-source state-of-the-art visual-inertial SLAM, SVIn, by integrating water-pressure and DVL measurements into a tightly-coupled nonlinear optimization formulation resulting in improved estimates. 
To leverage the structural information of recently available 3D acoustic sonars while maintaining computational efficiency, we introduce a hierarchical architecture that incorporates estimates from a decoupled front-end via KISS-ICP, treated as relative pose factors within the factor graph.
The proposed framework is validated through field deployments in a shipwreck and coral reef, demonstrating drift mitigation and global consistency. 
To support further research in underwater multi-modal fusion, the code and the datasets can be found on project website \url{https://cmburgul.github.io/projects/svin+/} \url{https://dartmouthrobotics.github.io/svin-plus}
\end{abstract}

\section{Introduction}
\label{sec:intro}

Reliable autonomy in underwater robotics needs robust Simultaneous Localization and Mapping (SLAM) in perceptually degraded GNSS-denied environments~\cite{mcconnell2022perception}, ranging from confined shipwrecks to dynamic coral reefs.
These underwater environments are characterized by many perceptual challenges, including light attenuation and limited visibility.

% As the physical robotics hardware becomes more accessible and embedded computing power increases, there is a more demand for autonomy and physical intelligence to work in  industrial automation, hazardous inspection, and the extreme environments, such as underwater caves—that remain inaccessible to humans.
%
% To enable this autonomy the robots must be capable of estimating their own motion state and understanding the surrounding environment. However, many of these environments are GPS-denied, preventing the robot from relying on external positioning systems.
%
% This challenge has led to the development of Simultaneous Localization and Mapping (SLAM) algorithms, in which a robot estimates its own trajectory while concurrently building a map of the environment by fusing measurements from onboard sensors.

Despite the plethora of SLAM approaches in the literature, robust state estimation in underwater domains remains an open problem.
Traditional Visual-Inertial Odometry (VIO) is prone to catastrophic failure due to feature tracking loss in turbid waters or motion-blur artifacts during high-dynamic maneuvers~\cite{JoshiIROS2019}.
Integrating water-pressure sensors (e.g.,~\cite{hu2021visual}) and Doppler Velocity Logs (DVLs) (e.g., ~\cite{ZhaoIROS23, VargasICRA21, xu2025aqua}) can mitigate water-column depth and velocity drift; however, these sensors provide no environmental context, leaving the system unable to perform the loop closures for global consistency.
Furthermore, although 2D multibeam sonars can offer some situational awareness, they suffer from geometric ambiguity in complex 3D structures~\cite{mcconnell2022perception}.

While a 3D acoustic sonar capable of generating volumetric point clouds has recently become available, and initial studies have begun to explore its potential, its integration into standard multi-sensor SLAM pipelines remains  underexplored. Although compact 3D sonars conceptually resemble terrestrial LiDARs, their point clouds are significantly sparser, have a lower update rate (5 Hz), and are heavily corrupted by acoustic multipath noise.
The few existing works have primarily addressed sonar-inertial configurations~\cite{archieri2026insso3d} or visual-inertial-acoustic fusion~\cite{pan2026viso,burgul2026arxiv}, without considering the full suite of sensors commonly used in Autonomous Underwater Vehicles (AUV).

\begin{figure}[t]
    \centering
    \begin{subfigure}{\linewidth}
        \centering
        \includegraphics[width=0.95\linewidth]{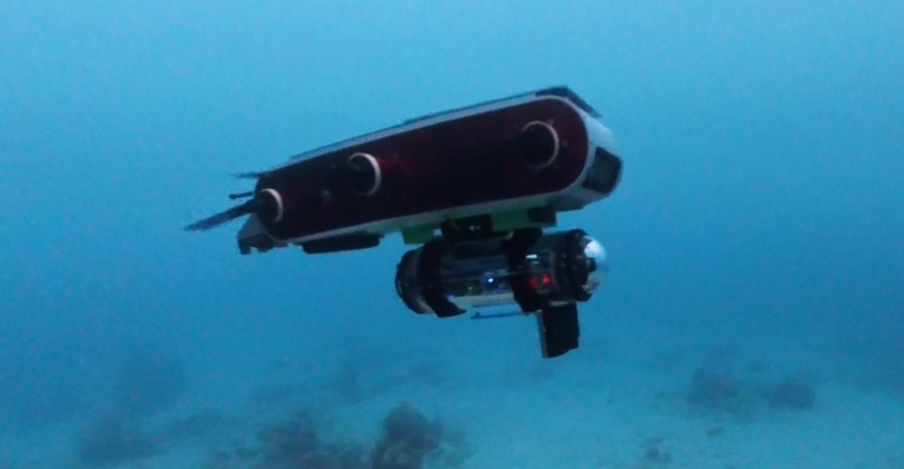}
        \vspace{-3mm}
        \label{fig:robot_setup_overview}
    \end{subfigure}
        % \vspace{-8mm}
    \begin{subfigure}{\linewidth}
        \centering
        \includegraphics[width=.95\linewidth]{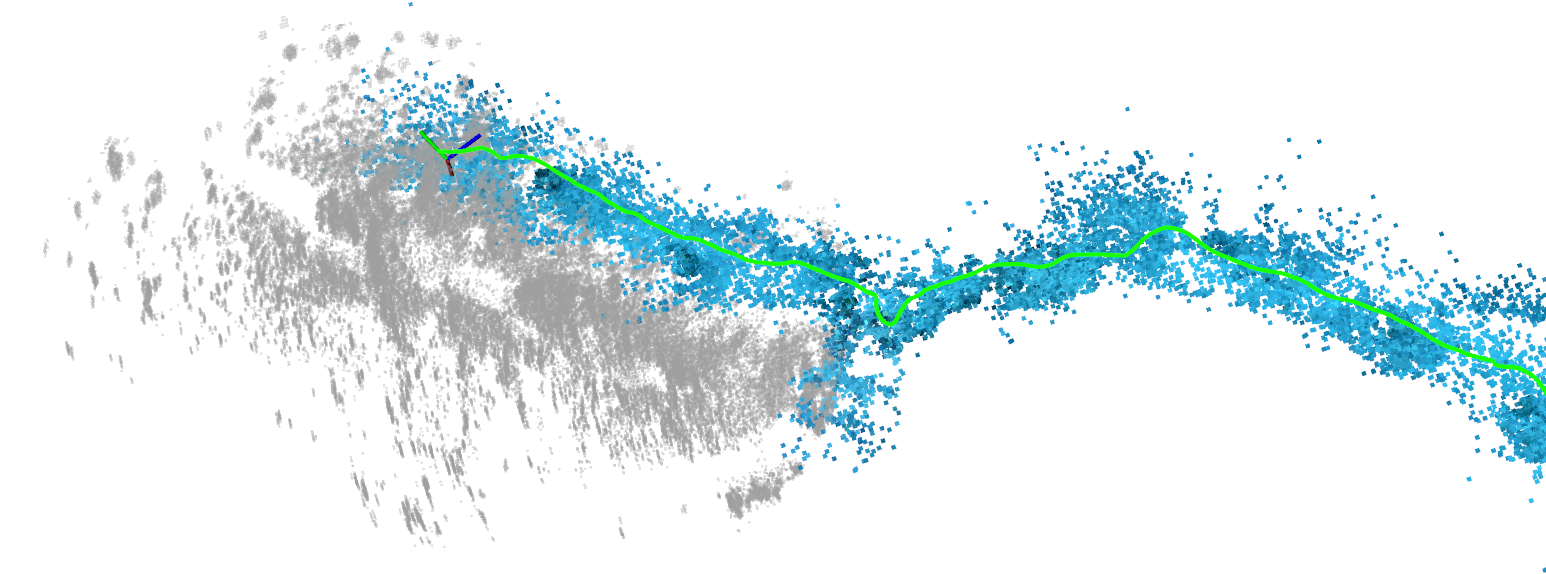}
        \label{fig:folkestonerviz}
    \end{subfigure}

    \caption{The Aqua2 robot equipped with multiple sensors during field deployment (top) and proposed \textit{SVIn+} multi-sensor odometry pose and trajectory (green), stereo point-cloud (blue) and the current local 3D sonar local map point-cloud (grey) on the Folkestone dataset (bottom).}
    \label{fig:robot_setup}
\end{figure}

    In this paper, we introduce a tightly-coupled multi-modal state estimation framework that achieves robust state estimation in challenging underwater environments. Our framework fuses inertial, camera vision, DVL, and water column depth from water-pressure sensor measurements into the factor graph. To provide structural consistency and mitigate trajectory drift, we integrate relative pose factors derived from a KISS-ICP front-end, which processes the point cloud from a recently released 3D acoustic sonar. 
    This decoupled approach abstracts 3D sonar data into relative 6D pose factors, preserving factor graph sparsity and enabling real-time operation on embedded hardware.

In summary, our main contributions are the following:
\begin{enumerate}
    \item Multi-modal state estimation framework, where vision, IMU, DVL, and water-pressure measurements are integrated in a tightly-coupled non-linear optimization for robust state estimation.
    \item Hierarchical 3D sonar integration, with a decoupled front-end via KISS-ICP for computational efficiency.
    \item Real-world validation in complex environments including shipwrecks and coral reefs, with the release of the associated code and datasets to the community.
\end{enumerate}

\section{Related Work}
\label{sec:related}

% \subsection{Underwater Multi-Sensor Navigation System}

Robust localization and mapping in underwater environments remains a major challenge due to the absence of GNSS signals forcing underwater vehicles to rely heavily on onboard sensing and dead reckoning, which leads to drift accumulation over time \cite{kinsey2006survey}. Furthermore, optical sensing is degraded by weak illumination, backscatter and turbidity \cite{yang2017acoustic}, while acoustic sensing suffers from multipath effects \cite{blondel1997handbook} and limited resolution \cite{negahdaripour20133}. As a result, multi-sensor integration is essential for achieving reliable, long-duration underwater SLAM.

Early approaches fuse Inertial Measurement Unit (IMU), Doppler Velocity Logs (DVL), depth, and acoustic positioning measurements within an Extended Kalman Filter (EKF) framework to reduce and manage uncertainties from heterogeneous sensors \cite{kinsey2006survey}.
While effective at managing short-term uncertainty, these systems lacked the dense environmental context needed to constrain long-term drift.

To achieve higher local accuracy, visual-inertial odometry (VIO) emerged as a dominant paradigm. 
Early vision-based underwater localization and mapping methods \cite{pizarro2004large} are primarily offline and rely on structure-from-motion (SfM) and bundle adjustment.
Subsequently, research shifted toward online, loosely coupled SLAM solutions. 
\cite{kim2013real} introduce a loosely coupled SLAM framework for underwater shipwreck inspection using a bag-of-words place recognition module and factor graph optimization.
\cite{hidalgo2018monocular} investigated the feasibility of performing underwater SLAM using ORB-SLAM \cite{mur2015orb}.
\cite{chavez2019adaptive,VargasICRA21,XuIROS21} are loosely coupled systems that integrate DVL with a visual–inertial SLAM framework.

More recently, researchers have moved toward tightly coupled visual–inertial systems for higher localization accuracy and estimator consistency.
\cite{ZhaoIROS23} adopts a Multi-State Constraint Kalman Filter (MSCKF)-based SLAM framework for tightly coupled visual–DVL multi-sensor fusion.
\cite{leonardi2023uvs,song2024turtlmap} use cameras as the primary sensing modality to achieve low-cost and dense reconstruction. 
Similarly, \cite{ding2025underwater} focuses on mitigating perception degradation through multi-sensor fusion.
AquaSLAM \cite{xu2025aqua} proposes an online SLAM framework that jointly estimates the camera–DVL extrinsic transformation within the state estimation process.
However, vision-based approaches remain fundamentally limited by restricted field of view, light attenuation, backscatter, and turbidity, and therefore typically operate only in relatively clear and well-illuminated environments.

To address these challenges, researchers have also adopted sonar as a primary perception modality. Sonar-based, vision-free SLAM methods \cite{fallon2013relocating, jie2018pose, 7989567, wang2022virtual, archieri20253dssdf} typically construct 2D maps under planar motion assumptions and are constrained by relatively low spatial resolution compared to optical sensing.
There are also methods using a mechanical sonar \cite{RahmanIJRR2022} or dual sonars \cite{chen20153d,mallios2017underwater,mcconnell2024large} for 3D mapping.
SVIn \cite{RahmanIJRR2022} implements a tightly coupled hybrid framework based on OKVIS \cite{LeuteneggerIJRR15}. While the incorporation of acoustic sensing improves robustness, the relatively low spatial resolution and low update frequency of the sonar measurements limit their ability to support dense reconstruction.

With the recent development of compact 3D sonar sensors, lightweight AUV systems can acquire denser point clouds at relatively higher update rates. There emerges hybrid approaches that combine the 3D sonar with the visual-inertial SLAM pipeline~\cite{burgul2026arxiv,pan2026viso,archieri2026insso3d}.
\cite{burgul2026arxiv} presents a loosely coupled fusion of visual and 3D sonar measurements for long-distance cave mapping.
\cite{archieri2026insso3d} adopts a loosely coupled framework that integrates 3D sonar with an inertial navigation system (INS) composed of a DVL, Attitude Heading Reference System(AHRS), and water-pressure sensor for pose estimation and mapping.
In contrast, \cite{pan2026viso} proposes a tightly coupled visual–inertial–sonar SLAM framework without DVL integration, and the system is evaluated only in a relatively small-scale ($6\,\mathrm{m} \times 7\,\mathrm{m}$) environment.
% \textit{3D Sonar based approaches:} The recent emergence of compact 3D sonars offers a new opportunity to capture dense geometry in turbid waters, overcoming the elevation ambiguity of imaging sonars and the slow scan rates of mechanical sonars. This is a nascent field; our preliminary work \cite{burgul2026arxiv} presented the first extrinsic calibration and loose fusion of a 3D sonar with VIO. Concurrent works have begun exploring this modality: Pan et al. \cite{pan2026viso} incorporated 3D sonar into a VINS framework for real-time dense mapping, while Archieri et al. \cite{archieri2026insso3d} explored fusing 3D sonar with inertial systems for large-scale SLAM.

Overall, tightly coupled multi-sensor fusion state-estimation framework for robots equipped with 3D sonar remains an open problem. A reliable and robust long-term multi-sensor state estimation solution that enables dense and accurate reconstruction in underwater environments, particularly in murky water, remains a challenge. 
In this paper, we address this gap by proposing a partially tightly coupled visual, sonar, DVL, water pressure framework that jointly optimizes measurements from multiple modalities including sonar, camera, water-pressure and DVL to address this gap.
% To the best of our knowledge, no unified framework currently exists that tightly fuses **Visual, Inertial, Pressure, DVL, and 3D Sonar** measurements into a single holistic SLAM factor graph. This proposal aims to bridge this gap, leveraging the complementary strengths of all five modalities to achieve robust estimation and dense mapping in challenging underwater environments.

\section{Preliminaries}

% Explain the state of SVIn

% \begin{align}
%     \mathbf{x}_k &= \begin{bmatrix} \mathbf{x}_{I_k}^\top & \mathbf{x}_{C_k}^\top  \end{bmatrix}^\top
% \end{align}\textbf{}

 % \subsection{Notation and Definitions}
This section introduces the notation and coordinate frames adopted in the paper.
The global reference (world) frame is denoted by ${G}$. The IMU frame is denoted by ${I}$, the camera frame by ${C}$, the sonar frame by ${L}$, the Doppler Velocity Log (DVL) frame by ${V}$, and the water-pressure sensor frame by ${D}$.
All sensor frames are rigidly attached to the robot body, and their extrinsic transformations are assumed to be known from calibration.
% 
% The state estimate at time $k$, denoted by $\mathbf{x}_k$, is defined as the rigid-body transformation ${}^{W}\mathbf{T}_{I}(t) \in SE(3)$, representing the pose of the IMU frame $\{I\}$ with respect to the global frame $\{G\}$.
The state vector consists of:
\begin{align}
    \mathbf{x}_k &= \begin{bmatrix}  \mathbf{x}_{I_k}^\top & \mathbf{x}_{H_k}^\top & \mathbf{x}_{Y_k}^\top \end{bmatrix}^\top \\
    \mathbf{x}_{{I_k}} &= \begin{bmatrix} \quat{I_k}{G}^\top & \p{I_k}{G}^\top & \vel{I_k}{G}^\top & \mathbf{b}_{g}^\top & \mathbf{b}_{a}^\top \end{bmatrix}^\top \label{eq:state_inertial}\\
    \mathbf{x}_{H_k} &= \begin{bmatrix} \mathbf{x}_{{I_{k-1}}} & \cdots & \mathbf{x}_{{I_{k-n}}} \end{bmatrix}^\top \\
    \mathbf{x}_{Y_k} &= \begin{bmatrix} \quat{I_{n-1}}{G}^\top & \p{I_{n\shortminus1}}{G}^\top & \cdots & \quat{I_{n-m}}{G}^\top & \p{I_{n\shortminus m}}{G}^\top \end{bmatrix}^\top
\end{align}
where $\mathbf{x}_{I_k}$ is the current body/IMU state, $\mathbf{x}_{H_k}$ are the $n$ historical non-keyframe states in the local optimization sliding window, $\mathbf{x}_{Y_k}$ are the $m$ keyframe states in the global frame $\{G\}$ for global pose-graph optimization\footnote{$\quat{I_k}{G}$ is the Hamilton unit quaternion corresponding to the rotation ${}^{G}_{I_k}\mathbf{R}$ from $\{I\}$ to  $\{G\}$,
$\p{I_k}{G}$ and $\vel{I_k}{G}$ are the position and velocity of $\{I\}$ in $\{G\}$, and $\mathbf{b}_{g}$ and $\mathbf{b}_{a}$ are the biases of the gyroscope and accelerometer, respectively.}.
Let $i \in \{C, L, V, D\}$ denote the sensor modality. The measurement generated by sensor $i$ at time $t$ is denoted by $\mathbf{z}_{i,t}$.
When the time index is clear from context, it is abbreviated as $\mathbf{z}_i$ for simplicity.

\begin{figure}[t]
    \centering
    \begin{subfigure}{\linewidth}
\centering\includegraphics[width=.85\linewidth]{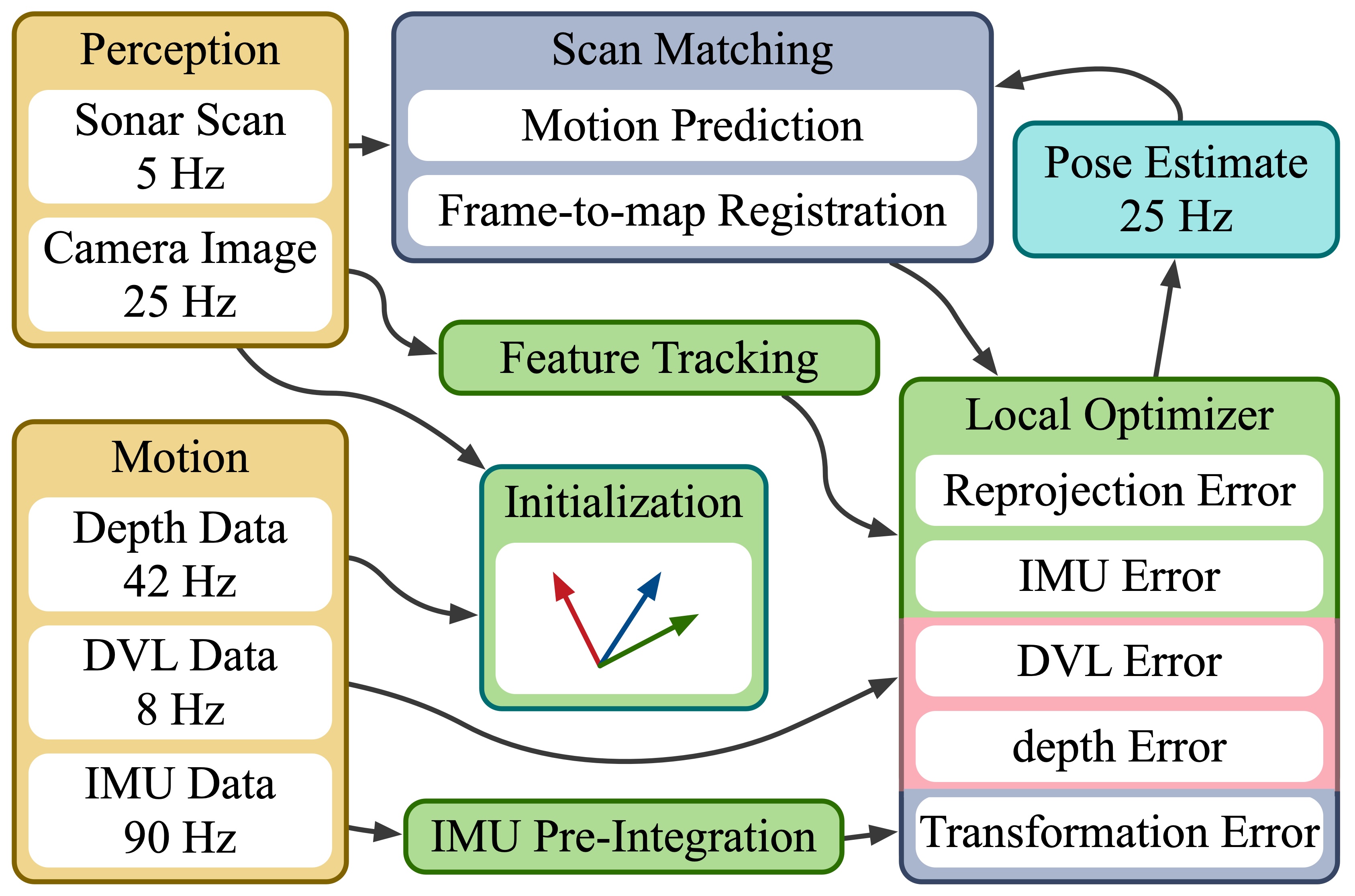}
    \label{fig:robot_setup_sensors}
    \end{subfigure}
    \caption{System pipeline of the proposed framework. Classical camera modules are highlighted in green, the sonar processing pipeline in gray, and marine-specific components in pink. This paper focuses on the integration of the local odometry optimization module.}
    \label{fig:pipeline}
\end{figure}

\section{Multi Sensor Odometry}
% The pipeline of the proposed SVIn+ framework is illustrated in Fig.~\ref{fig:pipeline}.
Our main contribution is the design and implementation of a local optimization module that provides high-frequency and reliable state estimation through the tightly coupled fusion of visual, inertial, water-pressure, DVL, and 3D sonar odometry measurements within a nonlinear optimization framework (Fig.~\ref{fig:pipeline}).
The local optimizer treats the state estimation problem as a nonlinear least-squares optimization over a sliding window of size $K$. Given $\mathbf{X} = \{\mathbf{x}_t,..., \mathbf{x}_{t+K}\}$, we formulate the optimization problem as:
\begin{align}
\mathbf{X} = \arg \min_{\mathbf{X}} J(\mathbf{X}).
\end{align}
The cost function $J(\mathbf{X})$ is defined as the sum of squared residuals corresponding to multiple sensing modalities, including the camera reprojection residual $\mathbf{e}_{\text{r}}$, inertial residual $\mathbf{e}_{\text{s}}$, DVL velocity residual $\mathbf{e}_{\text{v}}$, water-depth residual $\mathbf{e}_{\text{d}}$, and 3D sonar residual $\mathbf{e}_{\text{t}}$:

\begin{equation}
\begin{split}
    J(\mathbf{x}) = \frac{1}{2} \sum_{i}^{n} \sum_{k \in \mathcal{K}}^{K} \sum_{j \in \mathcal{J}(i,k)} {\mathbf{e}_{\text{r}}^{i,j,k}}^T \bm{\Omega}_{\text{r}}^{k} \mathbf{e}_{\text{r}}^{i,j,k} + \frac{1}{2} \sum_{k=1}^{K-1} {\mathbf{e}_{\text{s}}^{k}}^T \bm\Omega_{\text{s}}^{k} \mathbf{e}_{\text{s}}^{k}  \\
    + \frac{1}{2} \sum_{k}^{K-1} {\mathbf{e}^{k}_{\text{v}}}^T \bm\Omega^{k}_{\text{v}} \mathbf{e}^{k}_{\text{v}} 
     + \frac{1}{2} \sum_{k}^{K-1} \bm \Omega^{k}_{d} \left\| e_d^k \right\|^2   + \frac{1}{2} \sum_{k}^{K-1} {\mathbf{e}_{\text{L}}^{k}}^T \bm \Omega_{\text{L}} \mathbf{e}^{k}_{\text{L}}
\end{split}
\end{equation}
\noindent Here, $k$ denotes the frame index within the set of frames $\mathcal{K}$ contained in the sliding window, $i$ represents the camera index (allowing straightforward extension to an $n$ camera system), and $j\in\mathcal{J}$ indexes the visual landmarks observed in the $k$-th frame.
For a detailed derivation of the camera reprojection residual within a sliding-window nonlinear optimization framework, readers are referred to OKVIS \cite{LeuteneggerIJRR15}.
The inertial residual $\mathbf{e}_{\text{s}}$ is defined based on the IMU preintegration model, as described in \cite{forster2016manifold}.

The matrices $\bm\Omega_{\text{r}}^{k}, \bm\Omega_{\text{s}}^{k}, \bm\Omega_{\text{v}}^{k},$ and $\bm\Omega_{\text{t}}^{k}$ are the information matrices (inverse covariance matrices) for the visual, IMU, DVL, and 3D Sonar measurements at frame $k$, respectively. 
Note that while the visual, inertial, DVL, and sonar residuals are vectors, the depth error $e_d$ is a scalar, and is therefore weighted by the scalar $\Omega_{d}^{k}$.
The remainder of this section introduces the derivation of the DVL velocity residual, the depth residual, and the 3D sonar residual in detail.

\subsection{Water-Pressure Sensor Fusion}
The water-pressure sensor measures the water-pressure, which can be converted into water column depth ${z}_{d}$ relative to the water surface. 
The depth residual $e_{d}^{k}$ for a measurement ${z}_{d,k}$ at time step $k$ can be formulated similarly to \cite{ZhaoIROS23} as:
\begin{align}
    % z_{d,k} &= h_{d}(\mathbf{x}_{k}, D_{0}) + n_{d} \\
    e_{d}^{k} &= D_{0} - \mathbf{e}^\top_{3} (\R{G}{I_k}^\top \p{D}{I} + \p{I_k}{G})   
\end{align}   
where $\mathbf{e}^\top_{3} = \begin{bmatrix} 0 & 0 & 1 \end{bmatrix}$ extracts the Z-axis position in global frame, $\p{D}{I}$ is the translation from IMU to Water-Pressure Sensor. $D_{0}$ is the scalar depth of global frame $\{G\}$, computed using the initial depth measurement ${z}_{d,0}$ as $D_0 = {z}_{d,0} - \mathbf{e}^\top_{3} (\p{I_0}{G} + \R{I_0}{G}\p{D}{I})$.
Jacobians with respect to state can be defined as  $\frac{\partial \mathbf e_d}{\partial \p{I}{G}} = -\mathbf{e}^\top_{3}$, $\frac{\partial \mathbf e_d}{\partial \R{I}{G}} = \mathbf{e}^\top_{3} \R{I}{G} \lfloor  \p{D}{I} \times\rfloor$.
Where $\lfloor \cdot \times \rfloor$ denotes the skew-symmetric matrix operator mapping $\mathbf{v} \in \mathbb{R}^3$ to its corresponding skew-symmetric matrix in Lie algebra $\mathfrak{so}(3)$. Furthermore, given the short duration of the experiments compared to the tidal timescale, we assume a constant water surface level, neglecting the effects of tidal cycles and variations in ambient pressure.

\subsection{Doppler Velocity Log (DVL) Fusion}
The DVL provides linear velocity measurements $\mathbf{z}_{v}$ in the sensor frame. We relate these velocity to the system state velocity, following the approach in~\cite{ZhaoIROS23,xu2025aqua} to define the DVL residual $\mathbf{e_{v,k}}$ at time step $k$ as:
\begin{align}
    \mathbf{e_{v,k}} &= \mathbf{z}_{v,k} - \R{I}{V}(\R{G}{I_{k}} {}^{G}\mathbf{v}_{I_{k}} - \lfloor {}^{I}\omega_{k-1:k} \times\rfloor \p{V}{I})
\end{align}
where $\R{I}{V}$ and $\p{V}{I}$ are the extrinsic rotation from the IMU frame to DVL frame and position of DVL in IMU frame respectively. $\omega_{k-1:k}$ is the gyroscope angular velocity from $k-1$ to $k$ time steps.
The corresponding Jacobians with respect to state variables can be analytically derived as $\frac{\partial \mathbf e_v}{\partial \R{G}{I}} = - \R{I}{V} \lfloor \R{G}{I} {}^{G}\mathbf{v} \times\rfloor$, $\frac{\partial \mathbf e_v}{\partial {}^{G}\mathbf{v}} = - \R{I}{V} \R{G}{I}$.
We construct the information matrix $\bm\Omega_{v}$ using the velocity covariance $\bm\Sigma_v$ provided by the DVL sensor, allowing us to accurately weigh the measurement within the optimization framework.

\subsection{3D Sonar Fusion}
To process 3D Sonar measurements, we perform scan registration using KISS-ICP\cite{vizzo2023kiss}. 
Unlike the original implementation, which assumes a constant-velocity kinematic model for motion prediction, we use the pose estimate of previous step from SVIn+ as motion prior to register the current scan against a local map, yielding the sonar odometry ${}^{W}\mathbf{T}_{L} \in SE(3)$. Here, ${W}$ denotes the sonar odometry world frame, which is initialized to align with the first sonar frame ${S_0}$. This frame is distinct from the SVIn+ world frame ${G}$.

To account for the time offsets between the asynchronous sonar odometry and SVIn+ states, we perform $SE(3)$ interpolation on the bounding sonar odometry to derive the poses at the SVIn+ time steps $k-1$ and $k$.
By extracting these time-aligned odometry ${}^{W}_{L_{k-1}}\mathbf{T}$ and ${}^{W}_{L_{k}}\mathbf{T}$, we compute the relative transformation ${}^{L_{k-1}}_{L_{k}}\mathbf{T}_m$. Then, we construct the relative pose measurement defined by the rotation vector ${}^{L_{k-1}}\bs{\theta}_{L_{k}} \in \mathbb{R}^3$ and translation $\p{L_k}{L_{k-1}} \in \mathbb{R}^3$:
\begin{align}
    {}^{L_{k-1}}_{L_{k}}\mathbf{T}_m &= {}^{W}_{L_{k-1}}\mathbf{T}^{-1} {}^{W}_{L_{k}}\mathbf{T} \nonumber \\
    % \Delta \mathbf{T} &= \mathbf{T}_{WS_{k-1}}^{-1} \mathbf{T}_{WS_{k}} \nonumber \\ 
    &= \begin{bmatrix} \R{L_{k-1}}{W}^\top \R{L_k}{W} & \R{L_{k-1}}{W}^\top (\p{L_k}{W} - \p{L_{k-1}}{W}) \\ \mathbf{0}^\top & 1 \end{bmatrix} \\
    \p{L_k}{L_{k-1}} &= \R{L_{k-1}}{W}^\top (\p{L_k}{W} - \p{L_{k-1}}{W}) \\
    \bs{\theta}_{L,k-1:k} &= \log(\R{L_{k-1}}{W}^\top \R{L_k}{W})^\vee
\end{align}
where $\log(\cdot)^\vee$ extracts the rotation vector $ \bs{\theta} \in \mathbb{R}^3$ directly from the $SO(3)$ manifold.
%z
These components form the relative sonar measurement vector $\mathbf{z}_{L,k-1:k}$, subject to additive noise $\mathbf{n}_{L} \in \mathbb{R}^6$:
\begin{equation}
    \mathbf{z}_{L,k-1:k} = \begin{bmatrix} \p{L_k}{L_{k-1}} \\ {}^{L_{k-1}}\bs{\theta}_{L_{k}} \end{bmatrix} + \mathbf{n}_{L}
\end{equation}
To formulate the residual, we predict this relative transformation, denoted as ${}^{L_{k-1}}_{L_{k}}\mathbf{\hat{T}}$, using corresponding SVIn+ pose estimates ${}^{G}_{I_k-1}\mathbf{T}$ and ${}^{G}_{I_k}\mathbf{T}$ along with the IMU-Sonar extrinsic transformation:
\begin{align}
    {}^{L_{k-1}}_{L_{k}}\mathbf{\hat{T}} &= {}^{L}_{I}\mathbf{T} ~ {}^{G}_{I_{k-1}}\mathbf{T}^{-1} ~ {}^{G}_{I_k}\mathbf{T}  {}^{I}_{L}\mathbf{T} \\
    \phat{L_k}{L_{k-1}} &= \R{L}{I}^{\top} [\R{I_{k-1}}{G}^{\top}(\R{I_k}{G} \p{L}{I} + \p{I_k}{G} - \p{I_{k-1}}{G}) - \p{L}{I}] \\
    {}^{L_{k-1}}\bs{\hat{\theta}}_{L_k} &= \log(\R{L_k}{L_{k-1}})^\vee = \log(\R{L}{I}^{\top} \R{I_{k-1}}{G}^\top \R{I_k}{G} \R{L}{I})^\vee
\end{align}
Finally, the sonar relative pose residual $\mathbf{e}_{L,k-1,k}$ is computed as the difference between the interpolated measurement and the state prediction:  
\begin{align}
    \mathbf{e}_{L,k-1,k} &= \begin{bmatrix} \ptilde{L_k}{L_{k-1}} \\ {}^{L_{k-1}}\bs{\tilde{\theta}}_{L_k} \end{bmatrix} = \begin{bmatrix} \p{L_k}{L_{k-1}} - \phat{L_k}{L_{k-1}} \\ \log(\R{L_{k}}{L_{k-1}}^\top \Rhat{L_{k}}{L_{k-1}})^\vee \end{bmatrix}
\end{align}
The corresponding Jacobians with respect to state variables can be analytically derived as $\frac{\partial \mathbf \ptilde{L_{k-1}}{L_k} }{\partial \p{I_{k}}{G} } = \R{L}{I}^{\top} \R{I_{k-1}}{G}^{\top}$, $\frac{\partial \mathbf \ptilde{L_{k-1}}{L_k}}{\partial \R{I_{k}}{G} } = - \R{L}{I}^{\top} \R{I_{k-1}}{G}^{\top} \R{I_k}{G} \lfloor \p{L}{I} \rfloor$. 

% \subsection{Visual-Inertial Fusion}
% Write few lines of camera measurement and cite SVIn/OKVIS

% Write few lines of IMU measurement 
\section{Experimental Validation}
\label{sec:exp}

\begin{figure}
    \begin{subfigure}{\linewidth}
        \centering
        \includegraphics[width=0.9\linewidth]{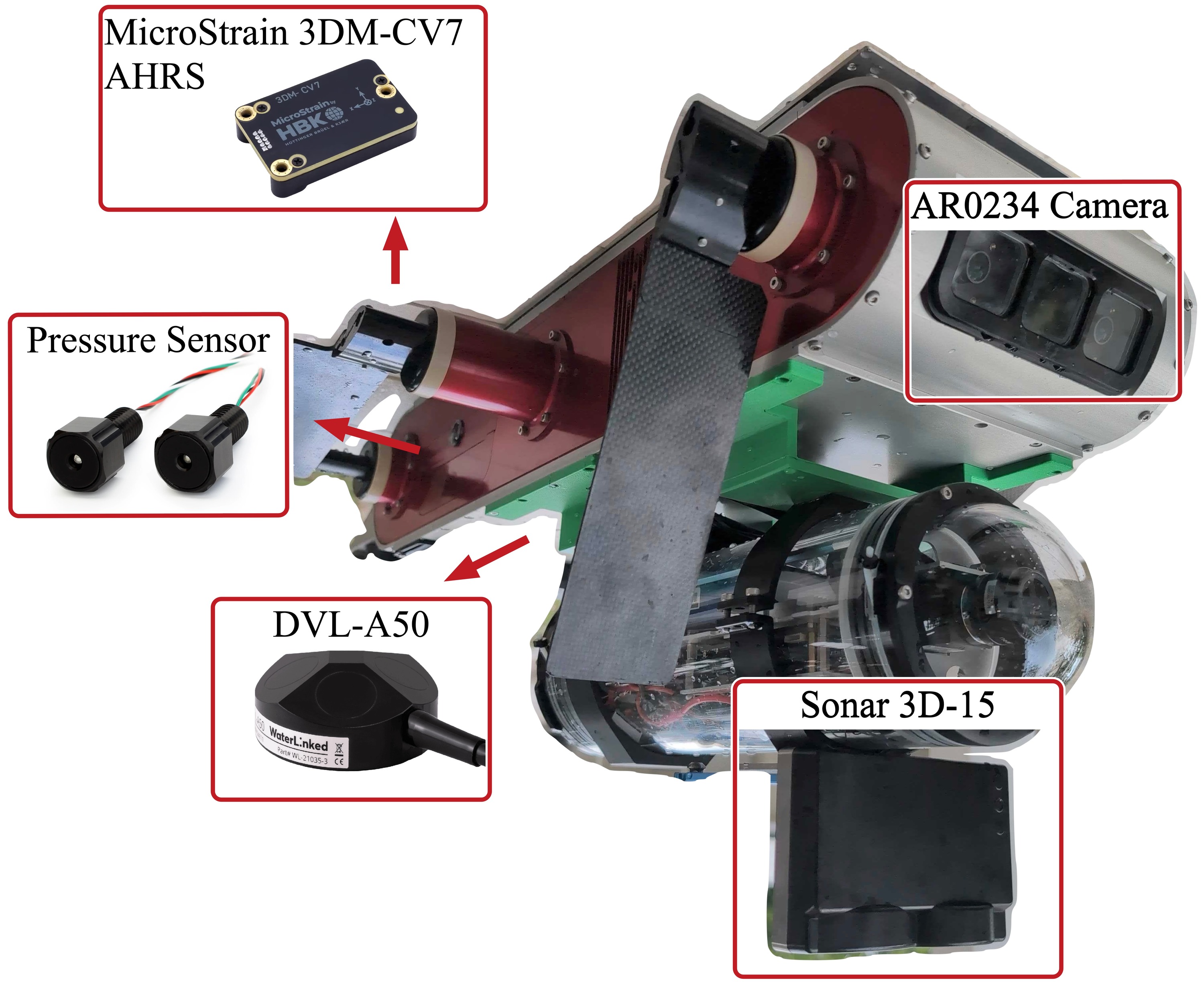}
    \end{subfigure}
    \caption{A detailed view of the onboard sensor configuration.}\label{fig:robot}
\end{figure}

To evaluate the performance and robustness of the proposed SVIn+, we collected a series of datasets in diverse and challenging underwater environments off the coast of Barbados. Our experiments aim to demonstrate that the integration of 3D sonar constraints into a tightly-coupled visual-inertial-DVL estimator mitigates trajectory drift by leveraging acoustic structural context. In the following subsections, we detail our hardware configuration and extrinsic calibration procedures, describe the characteristics of our shipwreck and coral reef datasets, and provide a quantitative comparison against state-of-the-art baselines.

% In this section, we validate SVIn+ on real-world datasets collected in Barbados using an 
\subsection{Experimental Setup}

We utilize the Aqua2 AUV from independent robotics equipped with a MicroStrain 3DM-CV7 AHRS, two AR0234 global shutter cameras in stereo configuration with hardware synchronization, a Water Linked DVL-A50 Doppler Velocity Log, a Bar30 pressure sensor, and a Water Linked 3D-15 sonar (Fig.~\ref{fig:robot}).

Accurately knowing the extrinsic transformation across this multi-sensor suite is critical for robust state estimation. 
The transformation between Camera $\{C\}$ and IMU $\{I\}$ reference frames is calibrated using Kalibr \cite{Furgale2013kalibr}.
The rigid extrinsic transformations from IMU to DVL $\{V\}$ and Water-Pressure Sensor $\{D\}$ frames are provided by the manufacturer of the Aqua2 robot.
Finally, the relative rotation and translation between 3D Sonar $\{L\}$ and Camera  $\{C\}$ are determined following the approach in \cite{burgul2026arxiv}. We solve the Perspective-n-Point (PnP) problem using 3D-2D correspondences extracted from four cinder block targets placed underwater and visible in both the sonar point clouds and visual images.
All experiments are conducted on a workstation running Ubuntu 24.04, equipped with an Intel Core Ultra 7 155H CPU, 64~GB of RAM, and an NVIDIA RTX 5000 Ada Generation GPU. The pipeline was also successfully evaluated in real-time on a low-power laptop (Intel Core i7-10510U CPU, 16 GB of RAM, NVIDIA GTX 1050), demonstrating its suitability for deployment on resource-constrained AUV hardware.

\subsection{Dataset Description}
%To evaluate the performance of SVIn+ in realistic underwater environments, 
We collected three datasets in two distinct locations off the coast of Barbados using the Aqua2 platform. To the best of our knowledge, there are currently no publicly available underwater datasets that provide a synchronized suite comprising stereo vision, DVL, and 3D  sonar data in such complex environments. 
These sequences encompass a range of perceptual challenges, from confined structural interiors to expansive open-water reefs:
\begin{itemize}
    \item \textit{BajanQueen-01%\& BajanQueen-02
    } (Shipwreck Interior): This sequence consists of a moderate-distance trajectory $\sim 75$ meters over about $8$ minutes, within the confined, structured environment of a shipwreck. Both missions start and end at a fixed ArUco marker, visited a total of 3 times, providing a reliable reference to evaluate start-to-end translation and rotation drift.% 45m, 283s
    \item \textit{BajanQueen-02} (Shipwreck): A longer, $\sim 600$-meter trajectory over around $55$ minutes, which covers the entire shipwreck, testing robustness against significant depth variations and sparse visual features.% 1181s
    \item \textit{Folkestone-01} (Shallow Reef): conducted in a natural unstructured environment, the trajectory spans $\sim 280$ meters over $1252$ seconds, challenging the vision frontend with dynamic lighting despite a relatively flat trajectory.
\end{itemize}
\subsection{Evaluation of Return-to-Origin Trajectory}
We evaluate the accumulated trajectory drift using \textit{BajanQueen-01} sequence. 
Since global ground truth (e.g., GPS or long-baseline acoustic positioning) is unavailable in these unstructured underwater environments, we rely on a ``return-to-origin'' metric.
We compare our proposed \textit{SVIn+} framework against two baselines: \textit{SVIn2} visual-inertial framework and KISS-ICP  as \textit{(Sonar Odom)}.  Global loop closures were disabled to isolate the performance of the underlying continuous odometry, ensuring that comparisons were not confounded by global correction mechanisms. 
In addition to the online baselines, we utilize an offline Structure-from-Motion (SfM) pipeline~\cite{schonberger2016structure}. We performed an extensive parameter tuning to get the best results from each method for fair comparison.
% To ensure an accurate comparison, all estimated trajectories are spatially aligned to their initial poses using the EVO package \cite{grupp2017evo}. 

\begin{figure}
    \begin{subfigure}{\linewidth}
        \centering
        \includegraphics[trim=2cm 1cm 4cm 2cm, clip, width=.9\linewidth]{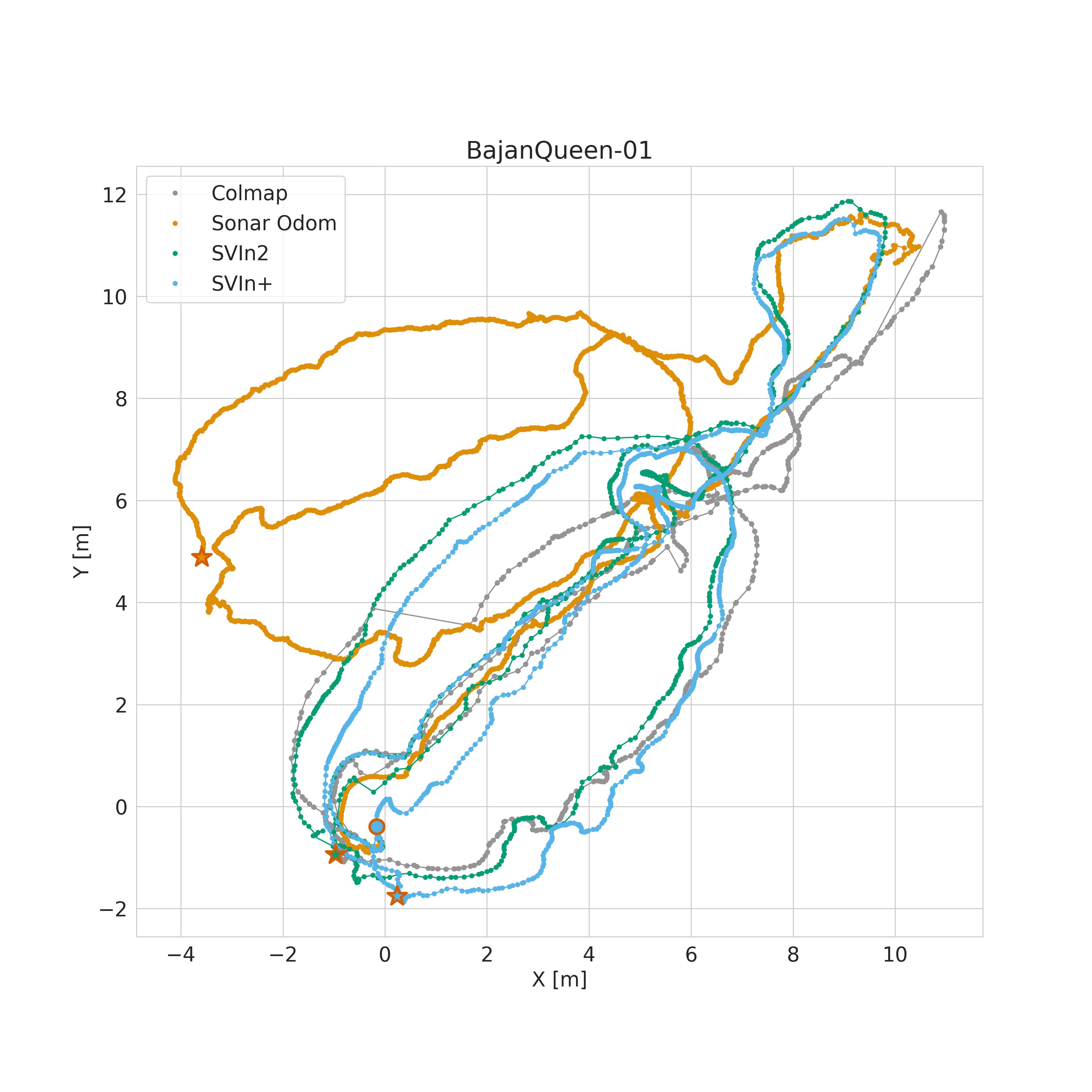}
        %{figures/bajan-queen-01/BajanQueen-01_Colmap_Sonar_Odom_Sonar_SVIn+_Prior_SVIn2_SVIn+_01.jpg}
        \vspace{-7mm}
    \end{subfigure}
    % \begin{subfigure}{\linewidth}
    %     \centering
    %     \includegraphics[trim=1cm 6cm 3cm 8cm, clip, width=.9\linewidth]{{figures/bajan-queen-01/BajanQueen-01_Sonar_Odom_SVIn2_SVIn+_12.jpg}}
    %     %{figures/bajan-queen-01/BajanQueen-01_Colmap_Sonar_Odom_Sonar_SVIn+_Prior_SVIn2_SVIn+_02.jpg}
    % \end{subfigure}
    \caption{Trajectory comparison between the proposed \textit{SVIn+} and the baseline \textit{SVIn2}, \textit{Sonar Odom} on the \textit{BajanQueen-01} dataset. %\textit{Top} shows the $x$-$y$ planar trajectory. 
    Evaluation of the start-to-end loop demonstrates the reduced accumulated drift achieved by the proposed multi-sensor fusion.}\label{fig:svin+_svin_comparison-01}
\end{figure}

Fig.~\ref{fig:svin+_svin_comparison-01} shows the trajectories over a complete start-to-end loop, with the initial and final positions denoted by a circle and a star, respectively. 
While Sonar Odom provides geometric context, it exhibited significantly higher drift due to the low-frequency nature of acoustic updates and noise (e.g., coming from multipath effects) in confined shipwreck compartments.
The offline SfM reference (Colmap) shows noticeable discontinuities where visual tracking failed due to environmental conditions.
%
% Qualitative analysis demonstrates that the \textit{SVIn+} trajectory terminates significantly closer to the true origin (anchored by the ArUco marker) compared to both baselines. 

\begin{figure}
    \centering
    \includegraphics[trim=0.0cm 8.0cm 0.0cm 5.2cm, clip, width=.9\linewidth]{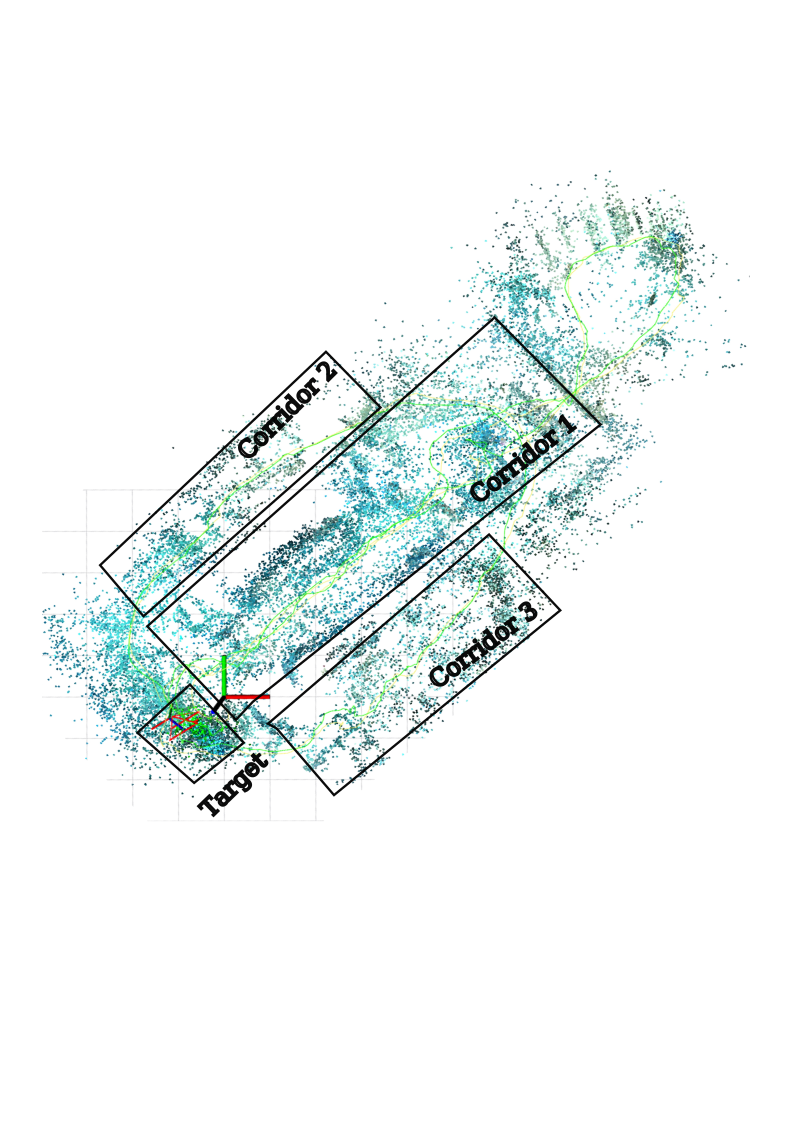}
    \caption{Visualization of the point-cloud from the \textit{SVIn+} on the \textit{BajanQueen-01} dataset.}
    \label{fig:svin+-point-cloud-bajan-01}
\end{figure}

\begin{figure*}
    \centering
    \includegraphics[trim=0.0cm 0.0cm 0.0cm 0.0cm, clip, width=0.8\linewidth]{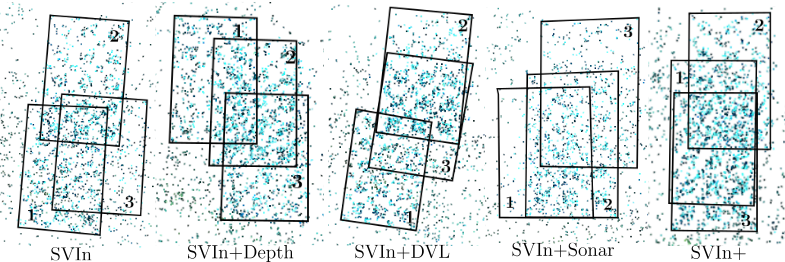}
    \caption{Visualization of accumulated drift utilizing an ArUco marker board. The return-to-origin error is quantified by evaluating the spatial deviation between the initial marker projection (1) and the projection recorded during the third revisit (3). Note that apparent variations in marker size are solely a result of perspective scaling due to varying distances from the camera.}
    \label{fig:reprojection-aruco}
\end{figure*}

SVIn+ overall maintains a consistent path throughout the dataset, where the start and end are close to each other, compared to SVIn2.
As shown qualitatively in Fig.~\ref{fig:reprojection-aruco}, the \textit{SVIn+} trajectory results in the ArUco marker board being closer compared to SVIn2 baselines. Quantifying the accumulated drift via the ArUco marker board reveals that the proposed \textit{SVIn+} achieves a return-to-origin error of just 0.52~m, representing a substantial improvement over the 0.81~m error yielded by \textit{SVIn2}. 
This lower drift is further evidenced by the resulting point cloud from SVIn+, which clearly shows three corridors on the left side of the shipwreck and the room on the right side (Fig.~\ref{fig:svin+-point-cloud-bajan-01}). 
These results highlight the effectiveness of the proposed multi-sensor fusion in bounding global drift and providing a coherent structural reconstruction.

\begin{figure*}[t]
    \begin{subfigure}{\textwidth}    
        \centering
        \includegraphics[trim=0.0cm 1.5cm 0.0cm 3.0cm, clip, width=0.85\textwidth]{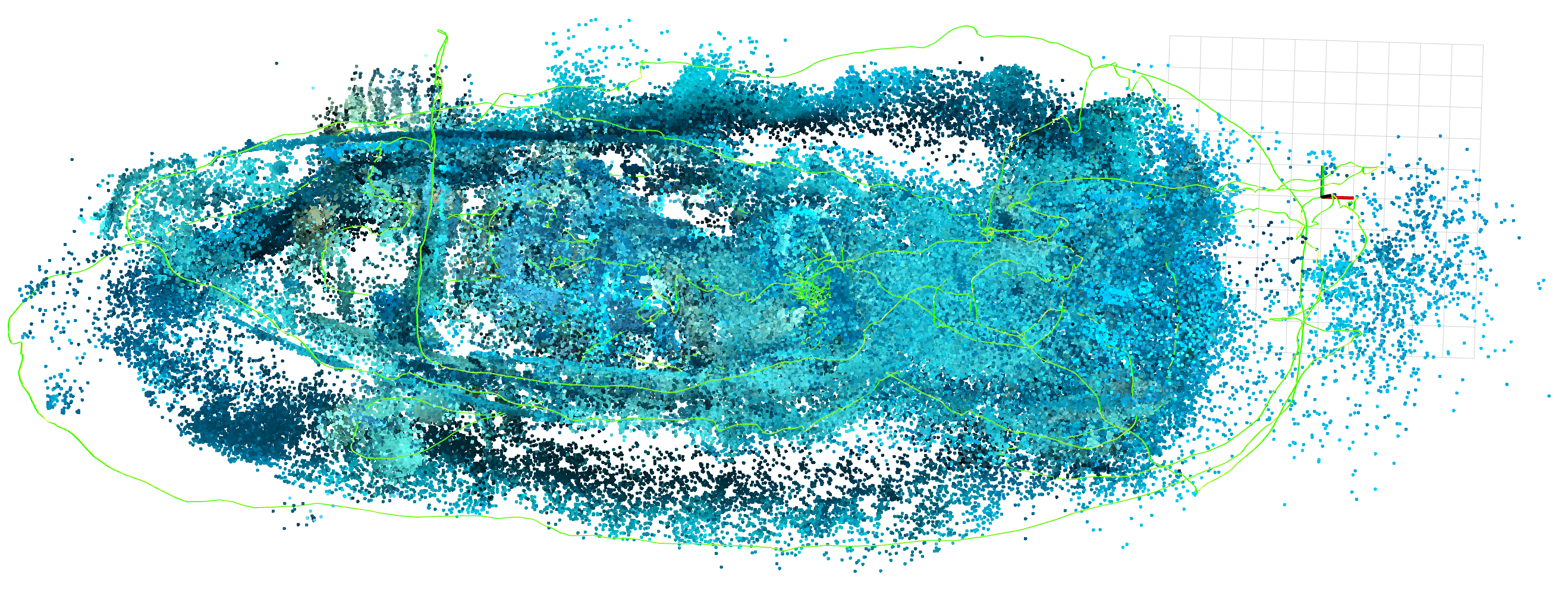}
    \end{subfigure}
    \begin{subfigure}{\textwidth}    
        \centering
        \includegraphics[trim=0.0cm 3.0cm 0.0cm 2.0cm, clip, width=\textwidth]{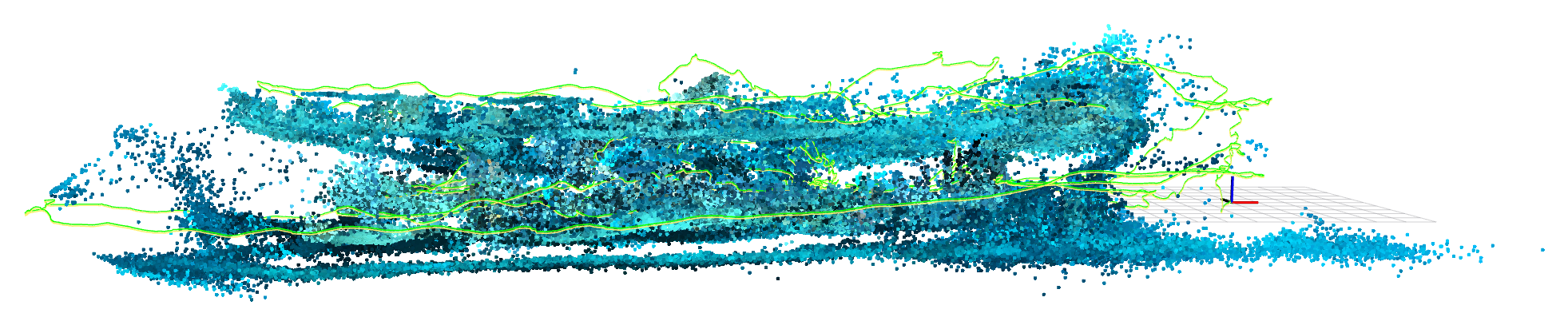}
    \end{subfigure}
    \caption{Visualization of the estimated trajectory and corresponding 3D point-cloud generated by \textit{SVIn+} Odometry on the \textit{BajanQueen-02} sequence. \textit{Top:} Top-down perspective of the mapped environment. \textit{Bottom:} Side view illustrating the depth profile as the Aqua2 robot navigates through the interior and around the exterior of the shipwreck structure.}
    \label{fig:svin+-top-bajan-03}
\end{figure*}

\subsection{Evaluation on Long-Duration Field Sequences}
To evaluate the long-term stability of \textit{SVIn+}, we further validate the system on two extended sequences characterized by significantly greater distances and durations.

\begin{figure}
    \centering
    ~~~\includegraphics[width=0.88\linewidth]{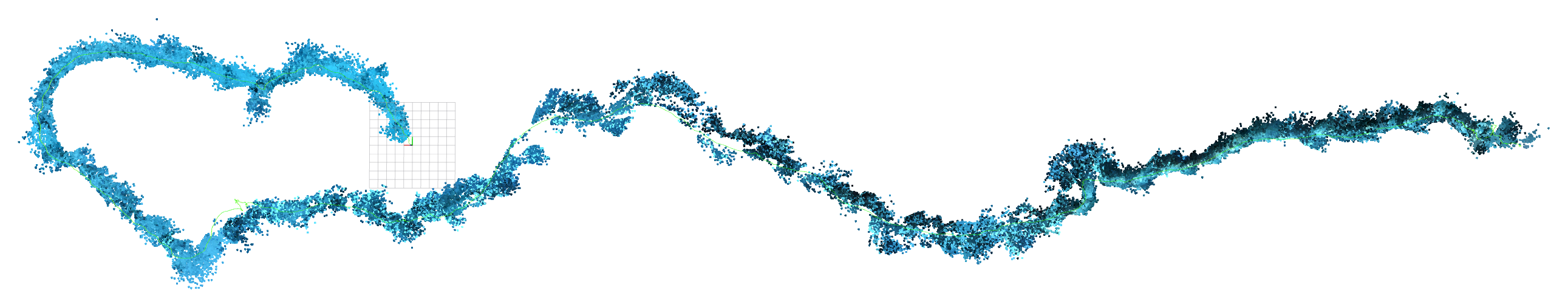}
    \includegraphics[width=\linewidth]{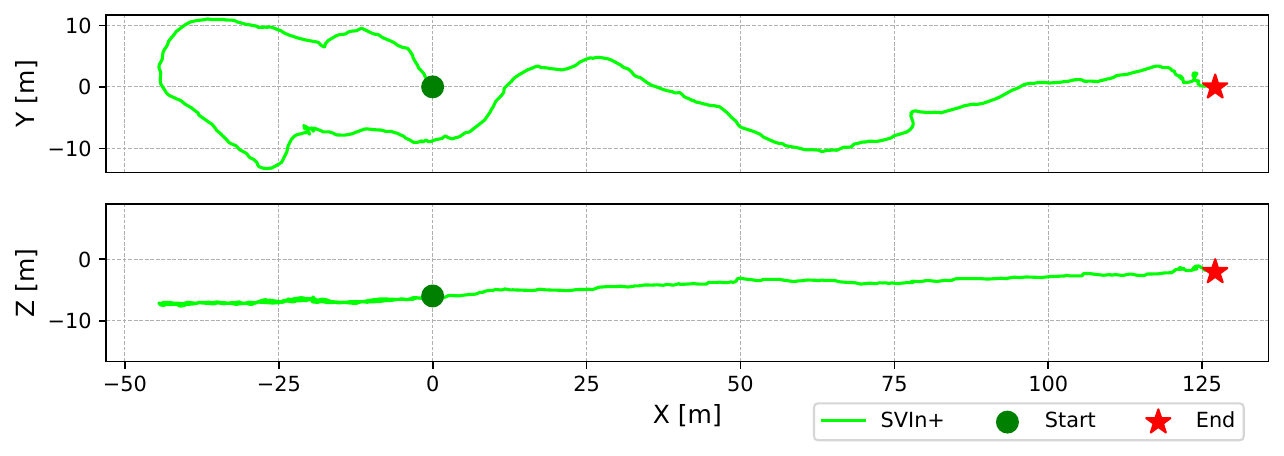}
    \caption{Top-down visualization of the trajectory with point-cloud from \textit{SVIn+} on the \textit{Folkestone} dataset (circle and star indicate start and end, respectively).}
    \label{fig:folkestone_visual}
\end{figure}

\fig{fig:svin+-top-bajan-03} illustrates the \textit{SVIn+}
 performance on the \textit{BajanQueen-02} shipwreck sequence, while
\fig{fig:folkestone_visual} demonstrates its capabilities on the \textit{Folkestone-01} coral reef sequence. Throughout these deployments, the estimator maintains a highly consistent global depth profile. The shipwreck side-view (Fig.~\ref{fig:svin+-top-bajan-03} bottom) demonstrates the ability to capture the shipwreck shape. The $y$-$z$ profile over the coral reef (Fig.~\ref{fig:folkestone_visual}, bottom) shows that the trajectory accurately captures the mission profile: the Aqua2 initially descends, executes a wide exploratory loop, and gradually ascends as it returns toward the shallows.  
Qualitatively, the visual coherence of the reconstructed shipwreck and reef features over their respective 215~m and 285~m transits demonstrates that the fusion of 3D sonar, water-pressure and DVL measurements provides robust constraints. This integration maintains a stable state estimate even in the presence of dynamic lighting and repetitive textural characteristics typical of shallow reefs.

% Need to explain in detail
% \begin{figure}
%     \begin{subfigure}{\linewidth}
%         \centering
%         \includegraphics[trim=1cm 2.2cm 3cm 1.5cm, clip, width=\linewidth]{figures/folkestone-corals/Folkestone_Sonar_Odom_Sonar_SVIn2_SVIn+_01.jpg}
%     \end{subfigure}
%     % \begin{subfigure}{\linewidth}
%     %     \centering
%     %     \includegraphics[trim=1.5cm 2cm 3cm 4.3cm, clip, width=\linewidth]{figures/folkestone-corals/Folkestone_Sonar_Odom_Sonar_SVIn2_SVIn+_02.jpg}
%     % \end{subfigure}
%     \caption{Trajectory comparison between the proposed \textit{SVIn+} and the baseline \textit{SVIn2}, \textit{Sonar Odom} on the \textit{BajanQueen-01} dataset. \textit{Top} shows the $x$-$y$ planar trajectory. Evaluation of the start-to-end loop demonstrates the reduced accumulated drift achieved by the proposed multi-sensor fusion.}\label{fig:svin+_svin_comparison-01}
% \end{figure}

% \begin{figure}
% \centering
% \includegraphics[width=\linewidth]{figures/bajan-queen-02/svin+_rviz.png}
% \caption{Visualization of the point-cloud from the \textit{SVIn+} on the \textit{BajanQueen-02} dataset.}
% \end{figure}

% \subsection{Mapping}

% \begin{figure}
%     \begin{subfigure}{\linewidth}
%         \centering
%         \includegraphics[width=\linewidth]{figures/folkestone-corals/svin+_rviz.png}
%     \end{subfigure}
%     \begin{subfigure}{\linewidth}
%         \centering
%         \includegraphics[width=\linewidth]{figures/folkestone-corals/svin2_rviz.png}
%     \end{subfigure}
%     \caption{[To DO:]Folkestone}
% \end{figure}

% \begin{figure}
%     \centering
%     \includegraphics[trim=0cm 0cm 0cm 0cm, clip, width=.7\linewidth]{figures/ablation/sq_aligned_comparison.png}
%     \caption{Trajectory comparison between SVIn2 with DVL integration and the standard SVIn2 baseline in pool dataset. The  return-to-origin evaluation highlights the smoother corners and improved trajectory continuity achieved by the addition of the DVL measurements.}
%     \label{fig:dvl_pool}
% \end{figure}

\begin{figure}
    \centering
    \includegraphics[width=\linewidth]{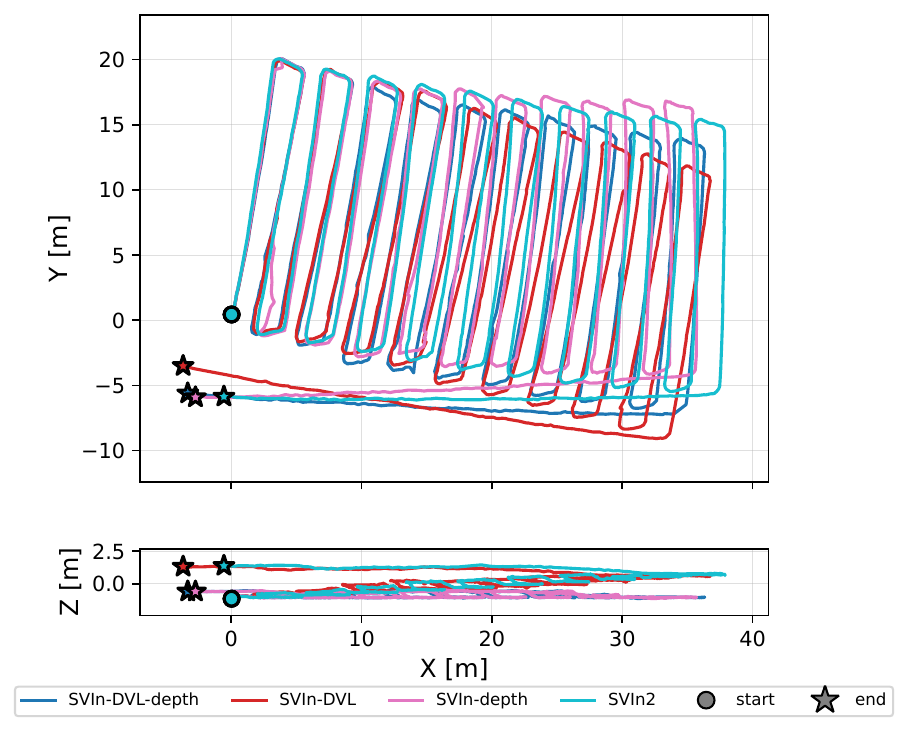}
    \caption{Comparison of estimated boustrophedon trajectories using the baseline SVIn2 framework against the proposed multi-sensor integrations (SVIn+Depth, SVIn+DVL, and SVIn+Depth+DVL).}
    \label{fig:buostrophedon}
\end{figure}

% \begin{figure}
%     \centering
%     \includegraphics[width=\linewidth]
%     {figures/ablation/bajan-01-depth-comparison.pdf}
%     % \includegraphics[trim=0cm 2.2cm 0cm 0cm, clip, width=\linewidth]{figures/ablation/depth_comparison.png}
%     \caption{Comparison of the estimated depth trajectories on \textit{BajanQueen-01} against the reference water-depth measurements. The results demonstrate that integrating water-pressure measurements into the \textit{SVIn2} framework yields a trajectory that tracks the reference depth profile significantly more closely than the standard \textit{SVIn2} baseline.}
%     \label{fig:depth}
% \end{figure}

\begin{table}[h]
    \centering
    \caption{Comparison of tag revisitation error, quantified by the average (and std) of the spatial deviations between reprojections of revisits of a fixed ArUco marker (as visualized in Fig.~\ref{fig:reprojection-aruco}).
    %the return-to-origin error, quantified by the spatial deviation between the initial and third ArUco marker reprojections (as visualized in Fig.~\ref{fig:reprojection-aruco}).
    }
    \label{tab:reprojection_distance}
    \begin{tabular}{lc}
        \toprule
        \textbf{Configuration} & \textbf{Reprojection Distance (m)} \\
        \midrule
        % old, measured from rviz
        % SVIn2      & 0.5228 \\
        % SVIn-depth & 0.528 \\
        % SVIn-dvl   & 1.242 \\
        % SVIn-Sonar & 0.400 \\
        % \textit{SVIn+}      & 0.524 \\
        % new, with fixed code and automated eval
        SVIn2      & 0.53 (0.11) \\
        SVIn2 (loop-closure) & 0.22 (0.11) \\
        SVIn-depth & 0.29 (0.11) \\
        SVIn-dvl   & 0.52 (0.23) \\
        SVIn-Sonar & 0.29 (0.15) \\
        \textit{SVIn+}      & \textbf{0.21 (0.08)} \\
        \bottomrule
    \end{tabular}
\end{table}

\begin{figure}
    \centering
    \includegraphics[trim=0.0cm 1cm 0 0, clip, width=0.87\linewidth]{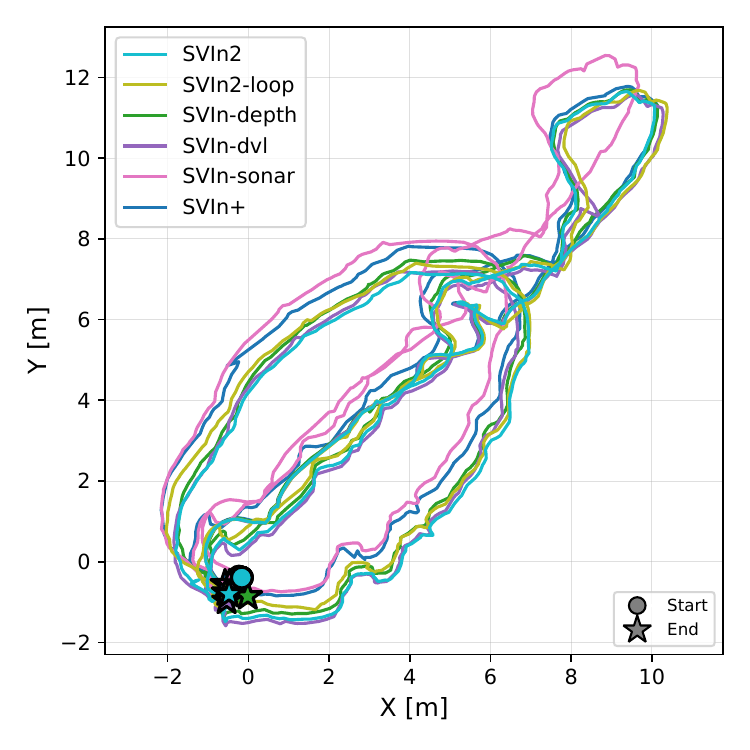}
    \caption{Trajectory comparison of baseline SVIn with adding different sensor modalities.}
    \label{fig:ablation_plot}
\end{figure}

\subsection{Ablation study}
To evaluate the individual contribution of each sensor modality, we performed an incremental ablation study. First, we performed an evaluation with the Aqua2 going over a shallow area at fixed depth and following a lawnmower pattern for qualitative assessment. Fig.~\ref{fig:buostrophedon} compares the estimated trajectories of baseline SVIn2 with DVL and Depth from water pressure sensor integration. The incorporation of DVL measurements effectively constrains drift in the $x-y$ plane, while depth integration bounds vertical $z$-axis error. The tightly-coupled fusion of both the sensors (SVIn+Depth+DVL) achieves the lowest overall positional drift from start to end.

To systematically isolate the contribution of each sensor modality, we evaluate five degraded configurations against our full system on the BajanQueen-01 dataset (Table~\ref{tab:reprojection_distance}, and Figs.~\ref{fig:reprojection-aruco},~\ref{fig:ablation_plot}), using the fixed marker in the environment. Starting from the visual-inertial baseline ($\text{SVIn2}$, mean error: $0.53\text{ m}$), the integration of the depth sensor alone ($\text{SVIn-depth}$) constrains vertical drift, reducing the mean deviation to $0.29\text{ m}$ ($45.3\%$ improvement). Fusing only the DVL ($\text{SVIn-dvl}$) provides bounded velocity constraints, yielding a mean error of $0.52\text{ m}$ with higher lateral consistency. The joint tightly-coupled integration of all modalities in $\text{SVIn+}$ yields the most accurate and geometrically consistent estimate ($0.21\text{ m}$, a $60.4\%$ overall error reduction).
Notably, the open-loop state estimation of $\text{SVIn+}$ ($0.21\text{ m}$ deviation) outperforms the baseline visual-inertial SLAM running with active loop closures ($\text{SVIn2-loop}$, $0.22\text{ m}$). This highlights that multi-modal fusion can match the global accuracy of loop-closed systems without the risk of false loop acceptance. 
% We then conducted a quantitative evaluation using \textit{BajanQueen-01} sequence, using the fixed marker in the environment. Starting from the baseline visual-inertial estimator (\textit{SVIn2}), we incrementally integrated depth, DVL, and finally 3D sonar constraints.
% As shown in Fig.~\ref{fig:depth}, the inclusion of pressure sensor data ensures the estimated depth profile follows the raw measurements more closely, effectively mitigating vertical drift.
%
% Furthermore, the integration of 3D sonar constraints ensures that the three successive passes over the marker are geometrically consistent and accurately aligned (Table~\ref{tab:reprojection_distance} and \fig{fig:reprojection-aruco}).
% Adding sensors shows an overall improvement with depth and sonar having the highest individual impact over the baseline.
% Notably, we could observe that the sonar odometry drifts, but was overall locally consistent and providing valuable spatial constraints.
% The best result is from integrating all the sensors together. We also ran SVIn2 with loop closure that got close results to our proposed method that did not include loop closure, demonstrating that the drift is minimal even without loop closure.
% Furthermore, the integration of 3D sonar constraints reduces the spatial dispersion of the estimated marker positions. 
As shown in \fig{fig:reprojection-aruco}, the reprojections from the three successive passes are concentrated within a noticeably tighter error radius, demonstrating enhanced geometric consistency.

%
% Notably, Table~\ref{tab:reprojection_distance} indicates that the baseline \textit{SVIn} integrated exclusively with sonar achieves a lower return-to-origin error than the complete \textit{SVIn+} framework for this specific trial. Rather than a shortcoming, this underscores the fundamental design objective of the proposed architecture: fault tolerance. Although no explicit sensor failures occurred during this dataset, and the inclusion of DVL measurements introduced larger positional errors, the fully integrated suite is critical for preventing catastrophic drift in the event of visual degeneracy or localized sensor dropouts. 
% This resilience is illustrated in Figure~\ref{fig:ablation_plot}, which depicts the return-to-origin trajectories as each modality is incrementally added to the \textit{SVIn2} baseline. The redundant constraints provided by Depth, DVL, and 3D sonar bound the overall estimation error. Ultimately, the full \textit{SVIn+} system successfully fuses this multi-sensor data to maintain stable, non-divergent state estimation even when individual modalities degrade or introduce noise. 

\begin{figure}
    \centering
    \includegraphics[trim= 0cm 0cm 0cm 0cm, clip, width=0.9\linewidth]{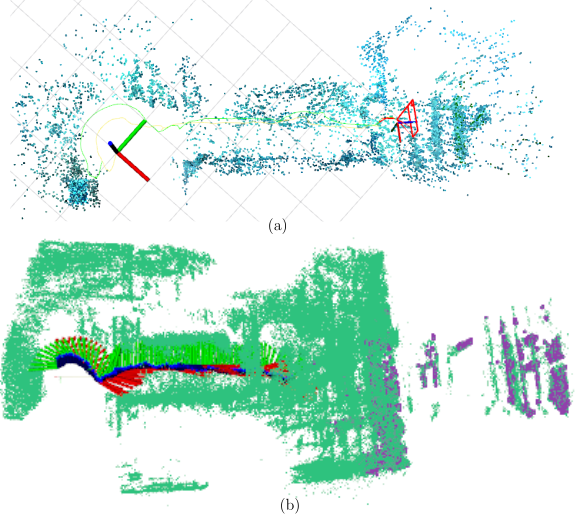}
    \caption{Comparison of (a) the 3D feature point cloud generated by the visual sensor and (b) the local map point cloud (green) constructed by registering 3D sonar point scans (purple) from the BajanQueen-01 dataset.}
    \label{fig:bajan-01-pc}
\end{figure}

Furthermore, Fig.~\ref{fig:bajan-01-pc} illustrates the dense map constructed from 3D Sonar scans compared to the 3D points extracted from visual measurements. This highlights the advantage of 3D Sonar: beyond enhancing localization, it generates rich environmental maps essential for subsequent path planning and execution of autonomous tasks.

%%%%%%%%%%%%%%%%%%%%%%%%%%%%%%%%%%%%%%%%%%%%%%%%%%%%%
%%%%%%%%%%%%%%%%%%%%%%%%%%%%%%%%%%%%%%%%%%%%%%%%%%%%%

\section{Lessons Learned and Future Work}
\begin{figure}
    \begin{subfigure}{.48\linewidth}
        \centering
        \includegraphics[angle=180, width=\linewidth]{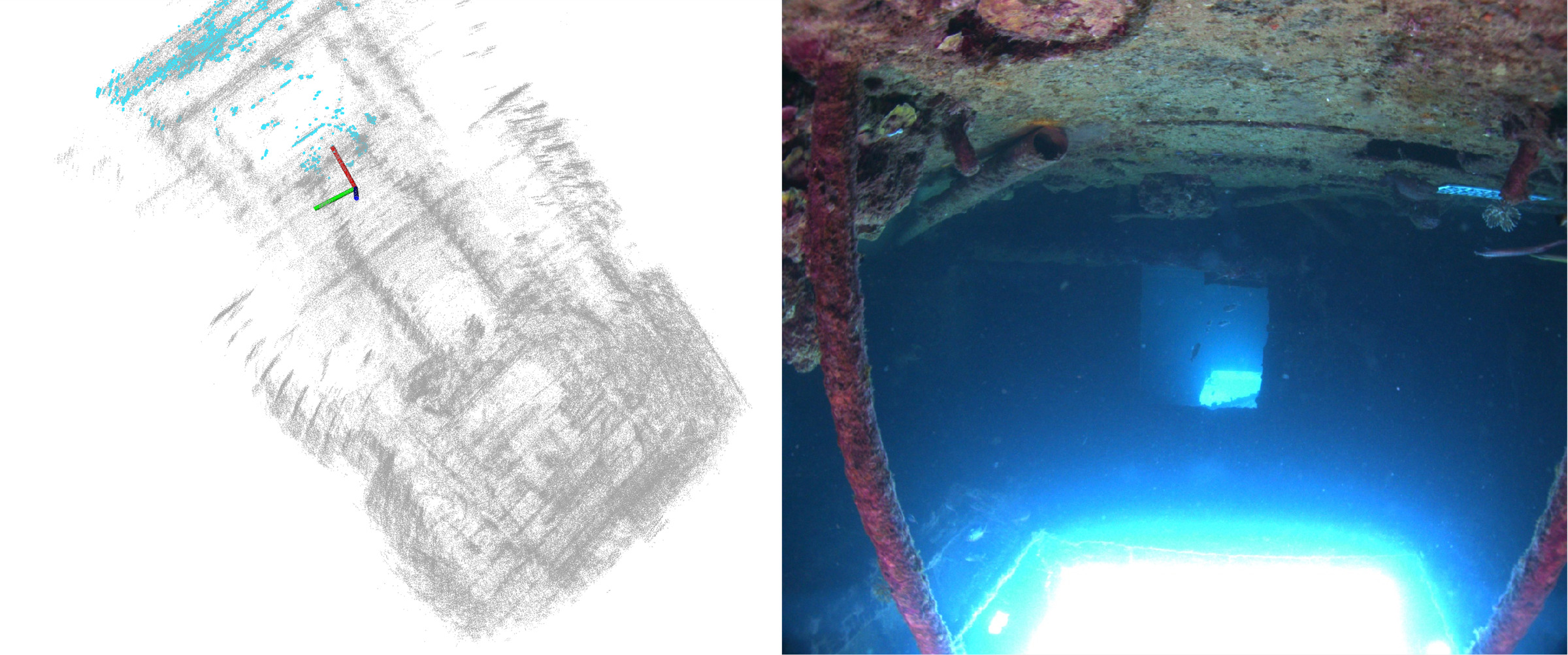}
        \caption{}
    \end{subfigure}\hfill
    \begin{subfigure}{.48\linewidth}
        \centering
        \includegraphics[angle=180, width=\linewidth]{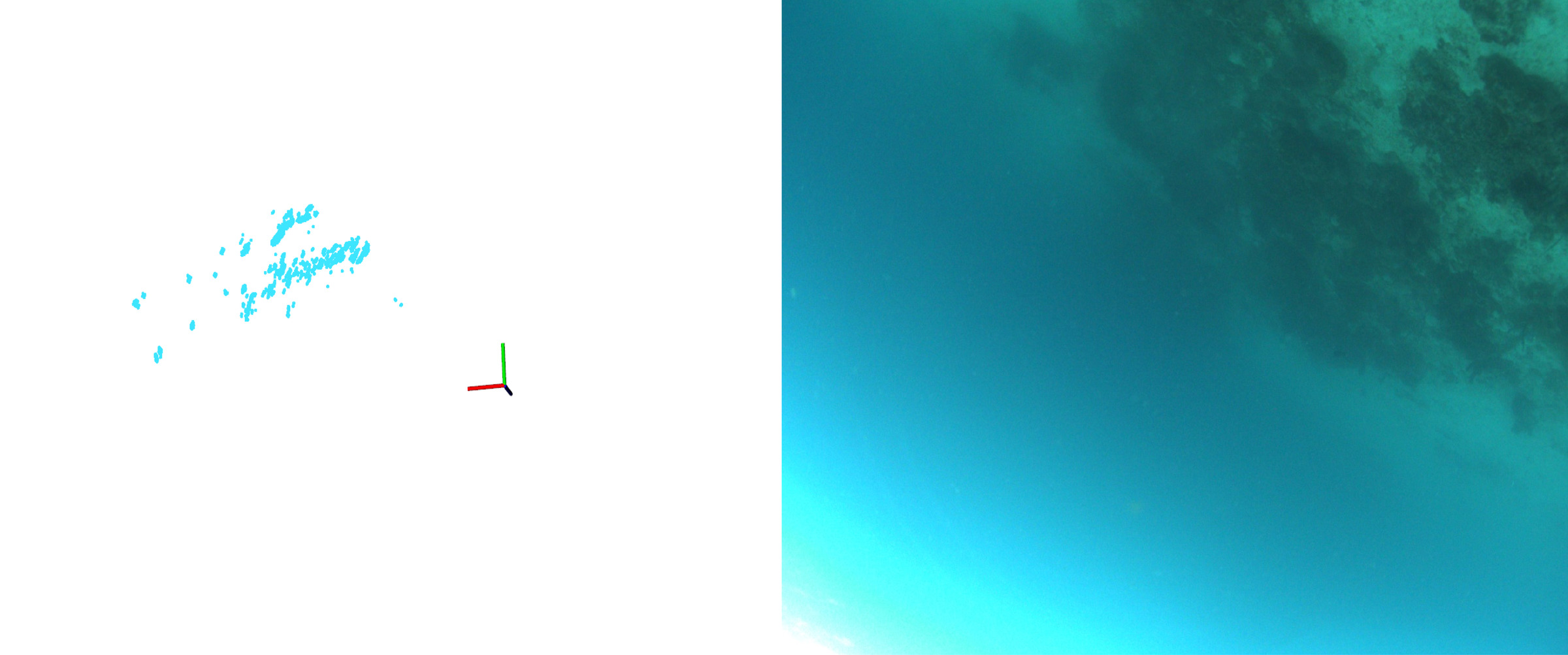}
        \caption{}
    \end{subfigure}
    \begin{subfigure}{.48\linewidth}
        \centering
        \includegraphics[angle=180, width=\linewidth]{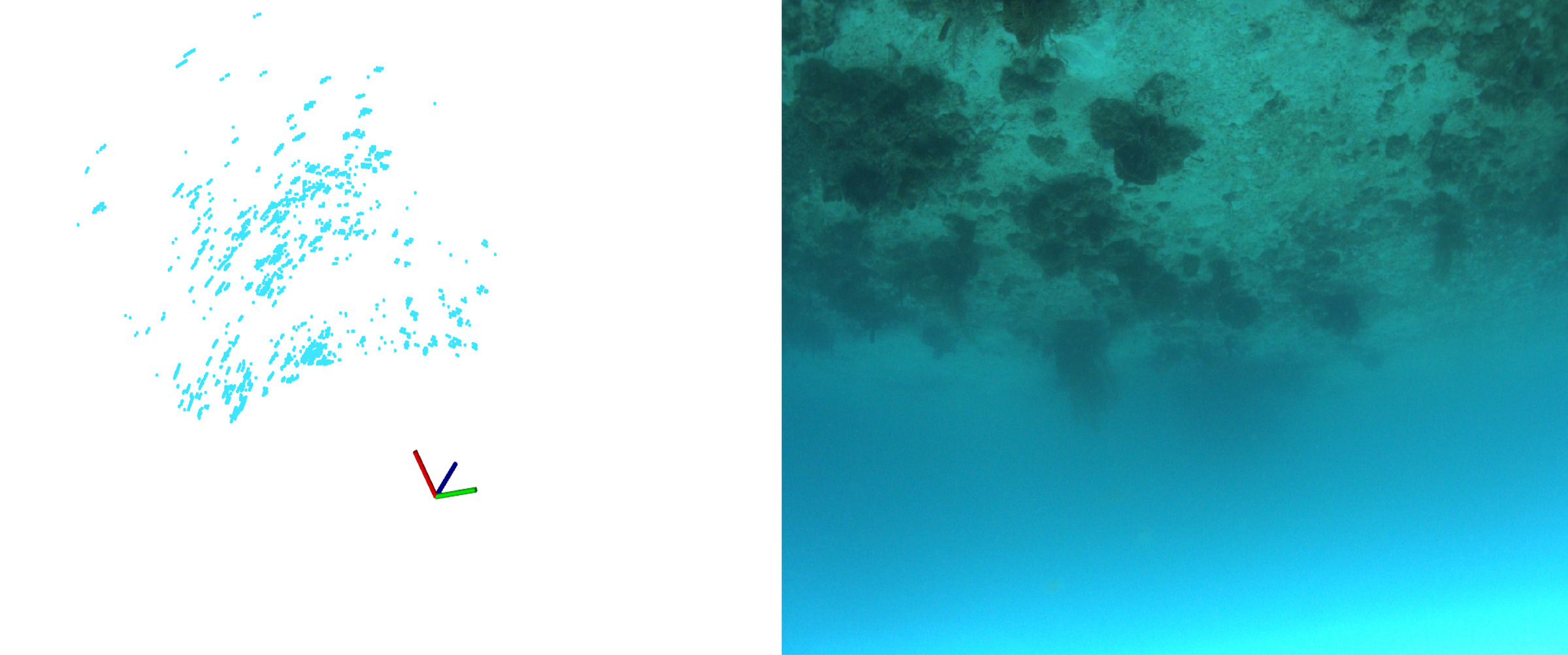}
        \caption{}
    \end{subfigure}\hfill
    \begin{subfigure}{.48\linewidth}
        \centering
        \includegraphics[angle=180, width=\linewidth]{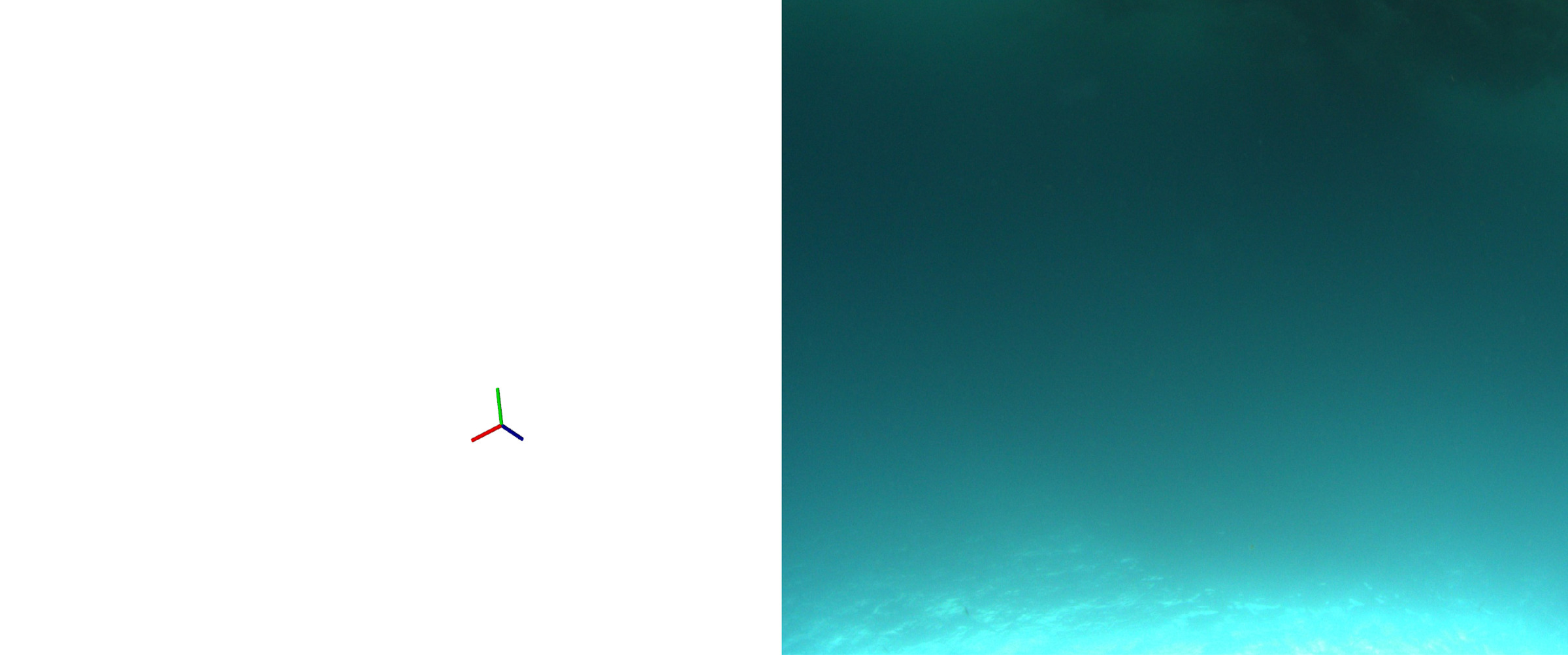}
        \caption{}
    \end{subfigure}
    \caption{Comparison of visual images and sonar point cloud scan \textit(blue) for a successful registration (a) and representative failure cases, where points are very sparse or non-present (b) (c) (d).}\label{fig:failure}
\end{figure}

The stability and drift reduction achieved by \textit{SVIn+} establish a robust foundation for several future research directions:
\newline
\textit{Robustness to sensor degeneracy}: While the current factor graph manages uncertainty, extreme conditions can render specific modalities completely unreliable (e.g., sparse 3D sonar registration failures, as depicted in \fig{fig:failure}). We plan to develop an adaptive switching mechanism for proactive failure detection and system resilience.
\newline
\textit{Global mapping and loop closure}: To transition from continuous odometry to a complete SLAM system for long-endurance missions, we will integrate multi-modal global place recognition. Leveraging both visual and acoustic signatures will ensure reliable re-localization and correct accumulated error.
\newline
\textit{Online spatial calibration}: Because unmodeled dynamic alignment variations (such as DVL misalignment) introduce systematic errors that offline calibration cannot fully resolve, we plan to implement an online extrinsic calibration backend to dynamically maximize trajectory accuracy.

\section{Conclusions}
\label{sec:conc}
In this paper, we presented SVIn+, a robust multi-modal framework for underwater state estimation that integrates measurements from 3D acoustic sonar, visual-inertial sensors, a water-pressure sensor and a Doppler Velocity Log (DVL).
To overcome the perceptual degradation common in underwater environments, such as light attenuation and turbidity, our tightly-coupled nonlinear optimization framework jointly optimizes these heterogeneous measurements within a local sliding window. To efficiently leverage the structural context of 3D sonars without compromising real-time performance, we introduced a hierarchical architecture that incorporates 6D relative pose factors from a decoupled KISS-ICP front-end.  
Extensive field experiments conducted with Aqua2 platform in complex environments, including a shipwreck and a coral reef, validate the effectiveness of our approach. Our evaluation demonstrates that the proposed multi-sensor fusion limits accumulated drift, outperforming the baseline SVIn2 framework and providing a robust foundation for future integration of global loop-closure and SLAM capabilities.
%
% For future work, we plan to extend this framework into a complete SLAM system by incorporating multi-modal global place recognition and loop closure modules.

% \section{Acknowledgments}
% \label{sec:ackn}

% \appendices
% \input{sections/9_appendix}
%%%%%%%%%%%%%%%%%%%%%%%%%%%%%%%%%%%
%%%%%%%%%%%%%%%%%%%%%%%%%%%%%%%%%%%
%%%%%%%%%%%%%%%%%%%%%%%%%%%%%%%%%%%

\bibliographystyle{IEEEtran}
% \bibliography{./IEEEabrv,bibliography/refs,bibliography/pubs, bibliography/sonar_cam_calibration}
\bibliography{./IEEEabrv,bibliography/refs,bibliography/afrl, bibliography/extra, bibliography/related, bibliography/rpng}

%%%%%%%%%%%%%%%%%%%%%%%%%%%%%%%%%%%%%%%%%%%%%%%%%%%%%%%%%%%%%%%%%%%%%%%%%%%%%%%%

\end{document}